\documentclass[conference]{IEEEtran}
\IEEEoverridecommandlockouts

\usepackage[compact]{titlesec}
\usepackage{cite}
\usepackage{amsmath,amssymb,amsfonts}
\usepackage{algorithmic}
\usepackage{graphicx}
\usepackage{textcomp}
\usepackage{xcolor}

\usepackage{algorithm}

\def\BibTeX{{\rm B\kern-.05em{\sc i\kern-.025em b}\kern-.08em
    T\kern-.1667em\lower.7ex\hbox{E}\kern-.125emX}}
\begin{document}

\title{Rethinking Adversarial Attacks in Reinforcement Learning from Policy Distribution Perspective\\
\thanks{Copyright 2025 IEEE. Published in ICASSP 2025 – 2025 IEEE International Conference on Acoustics, Speech and Signal Processing (ICASSP), scheduled for 6-11 April 2025 in Hyderabad, India. Personal use of this material is permitted. However, permission to reprint/republish this material for advertising or promotional purposes or for creating new collective works for resale or redistribution to servers or lists, or to reuse any copyrighted component of this work in other works, must be obtained from the IEEE. Contact: Manager, Copyrights and Permissions / IEEE Service Center / 445 Hoes Lane / P.O. Box 1331 / Piscataway, NJ 08855-1331, USA. Telephone: + Intl. 908-562-3966.}
}

\author{
 Tianyang Duan\textsuperscript{\rm 1},
    Zongyuan Zhang\textsuperscript{\rm 1},
    Zheng Lin\textsuperscript{\rm 2}, 
    Yue Gao\textsuperscript{\rm 3},
    Ling Xiong\textsuperscript{\rm 4}, \\
    Yong Cui\textsuperscript{\rm 5},
    Hongbin Liang\textsuperscript{\rm 6},
    Xianhao Chen\textsuperscript{\rm 2},
    Heming Cui\textsuperscript{\rm 1},
    Dong Huang\textsuperscript{\rm 1}\\
    \textsuperscript{\rm 1} Department of Computer Science, The University of Hong Kong, Hong Kong, China.\\
    \textsuperscript{\rm 2} Department of Electrical and Electronic Engineering, The University of Hong Kong, Hong Kong, China.\\
    \textsuperscript{\rm 3} School of Computer Science, Fudan University, Shanghai, China.\\
    \textsuperscript{\rm 4} School of Computer and Software Engineering, Xihua University, Chengdu, China. \\
    \textsuperscript{\rm 5} Department of Computer Science and Technology, Tsinghua University, Beijing, China.\\
    \textsuperscript{\rm 6} School of Transportation and Logistics, Southwest Jiaotong University, Chengdu, China. \\
}

\maketitle

\begin{abstract}
Deep Reinforcement Learning (DRL) suffers from uncertainties and inaccuracies in the observation signal  in real-world applications. Adversarial attack is an effective method for evaluating the robustness of DRL agents. However, existing attack methods targeting individual sampled actions have limited impacts on the overall policy distribution, particularly in continuous action spaces. To address these limitations, we propose the Distribution-Aware Projected Gradient Descent attack (DAPGD). DAPGD uses distribution similarity as the gradient perturbation input to attack the policy network, which leverages the entire policy distribution rather than relying on individual samples. We utilize the Bhattacharyya distance in DAPGD to measure policy similarity, enabling sensitive detection of subtle but critical differences between probability distributions. Our experiment results demonstrate that DAPGD achieves SOTA results compared to the baselines in three robot navigation tasks, achieving an average 22.03\% higher reward drop compared to the best baseline.
\end{abstract}


\section{Introduction}

Deep Reinforcement Learning (DRL) has exhibited exceptional performance across a spectrum of complex robotic control tasks~\cite{xu2021crpo,achiam2017constrained,yang2022cup}, successfully solving challenges such as robot navigation\cite{surmann2020deep,zhu2021deep,lin2023pushing}, obstacle avoidance\cite{wang2023safe,wang2023obstacle}, and robotic manipulation\cite{kilinc2022reinforcement,franceschetti2021robotic}. However, in real-world applications, observation signals are often subject to noise, latency, and inaccuracies\cite{ditzler2015learning,lin2024split}. Moreover, subtle variations in environmental dynamics, such as changes in surface friction or lighting conditions, can substantially degrade DRL performance\cite{padakandla2021survey}.

\begin{figure}[!t]
  \centering
    \includegraphics[width=0.93\linewidth]{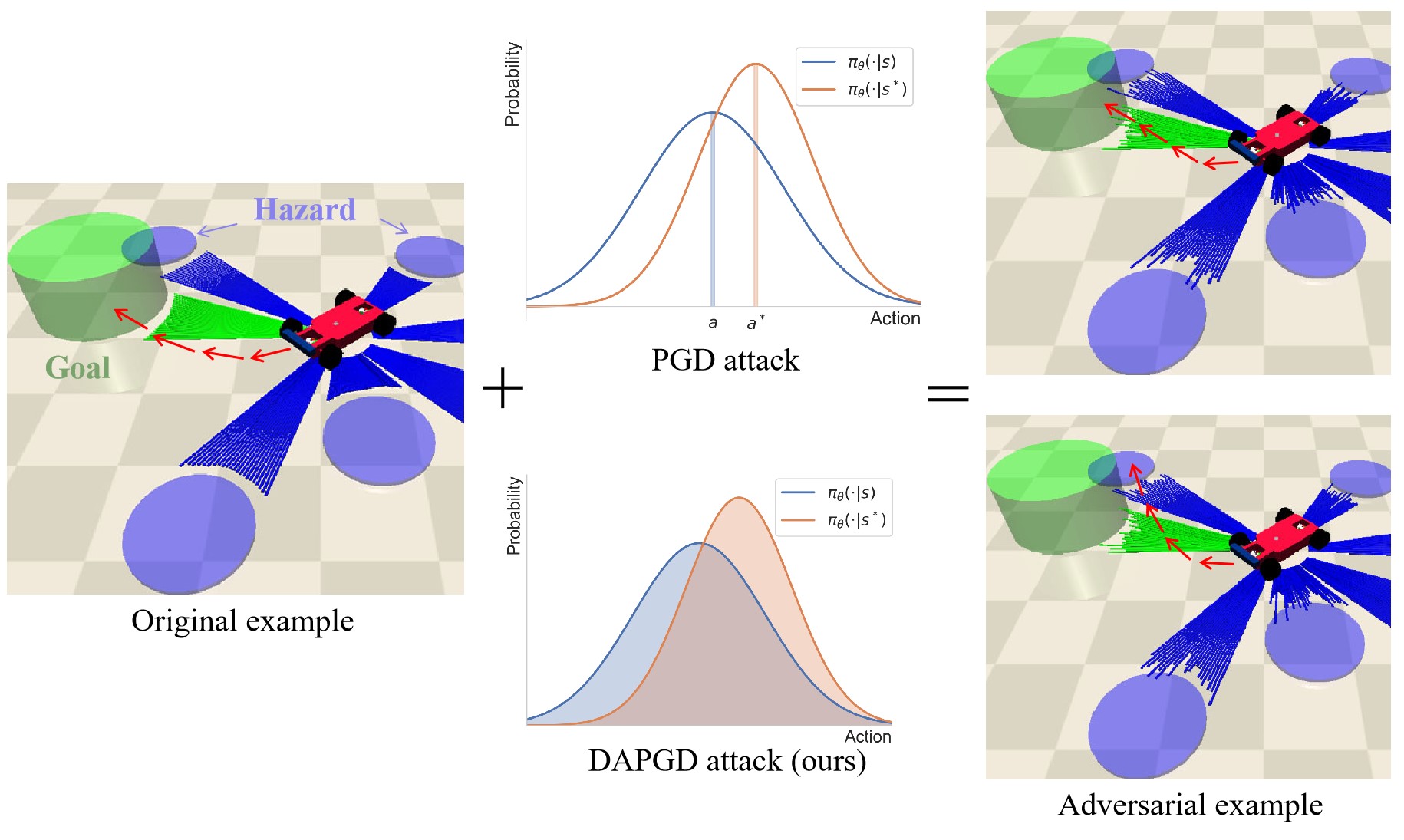}
      \vspace{-1.5ex}
    \caption{Two methods for generating adversarial examples in the Goal task. In this task, the agent needs to navigate around \textit{Hazards} and reach the \textit{Goal}. \textbf{Top}: Existing methods (e.g., PGD) sample from the policy and calculate the sign gradient of mean square error loss to attack. \textbf{Bottom}: Our method (DAPGD) directly utilizes the policy distribution similarity, which calculates the sign gradient of the  Bhattacharyya distance between policies to attack.}
    \vspace{-1ex}
    \label{fig:1}
\end{figure}

Adversarial attacks have emerged as an effective method for evaluating the robustness of DRL models\cite{xu2020adversarial,chakraborty2021survey}. By introducing carefully designed perturbations to input data, these attacks effectively reveal the vulnerabilities of DRL models under uncertainties and environmental changes\cite{pattanaik2017robust}. Training DRL agents in the presence of such adversarial scenarios has been shown to improve their robustness\cite{korkmaz2023adversarial}.
While current research on adversarial attacks primarily focuses on supervised learning tasks, such as image classification\cite{su2019one,modas2019sparsefool,pomponi2022pixle}, the application in the domain of DRL remains relatively unexplored. Existing efforts have concentrated on developing defense mechanisms compatible with DRL frameworks\cite{sun2020stealthy,zhang2021robust,oikarinen2021robust}, while little attention is paid to devising attack methods against DRL, particularly in high-dimensional continuous state and action spaces.

Unlike supervised learning~\cite{lin2024adaptsfl,hu2024accelerating,zhang2024satfed,lin2024efficient}, which involves deterministic input-output mappings~\cite{lin2024fedsn,zhang2024fedac,lin2024hierarchical}, DRL learns a policy function that maps state space to action space, typically modeled as a conditional probability distribution over possible actions\cite{wang2022deep}. As illustrated in the top of Figure~\ref{fig:1}, existing adversarial attack methods primarily target the output, which in DRL manifests as calculating adversarial perturbations based on sampled actions from the policy (or the action with the highest probability)\cite{huang2017adversarial,lin2017tactics}. However, perturbation attacks targeting individual actions may have a limited impact on the overall policy distribution, particularly in high-dimensional continuous state and action spaces. This is because adjacent actions within the action space can be selected to counteract the effects of attacks targeting specific actions. The inherent flexibility of continuous action spaces enables more robust adjustments in response to local perturbations. Moreover, existing methods often overlook the inherent randomness in action selection. Consequently, gradients of the loss function calculated from a single sampled action fail to capture the broader vulnerability of the entire policy distribution, thereby diminishing the effectiveness of the attacks.

To address these issues, we propose the \textit{Distribution-Aware Projected Gradient Descent (DAPGD)} attack. As shown in the bottom of Figure~\ref{fig:1}, our method utilizes distribution similarity, instead of sampled outcomes, as the input for gradient perturbation to attack the policy network. This approach capitalizes on the full information of the policy distribution rather than relying solely on the local characteristics of individual samples. Furthermore, this gradient can generate more consistent perturbation effects across the observation space, resulting in adversarial samples that more closely resemble real-world disturbances. This is particularly crucial for assessing DRL robustness in practical application scenarios. In our DAPGD attack method, we employ the Bhattacharyya distance as a measure of policy similarity. The Bhattacharyya distance\cite{kailath1967divergence} quantifies the overlap between two probability distributions and is sensitive to marginal changes, facilitating the detection of subtle but potentially critical differences in policies. Experimental results show that DAPGD outperforms seven state-of-the-art baselines across three robotic navigation tasks. DAPGD achieves an average reward reduction 18.67\% and 25.38\% higher than the best baseline when attacking benign and robust models, respectively.

\section{Related Work}
Adversarial attacks are predominantly applied to supervised learning tasks such as image recognition\cite{goodfellow2014explaining,madry2017towards, bu2023towards}. FGSM\cite{goodfellow2014explaining} is a basic attack method that generates adversarial samples by perturbing inputs based on the sign of gradients in a single step, evaluating the robustness of deep neural networks (DNNs)~\cite{peng2024sums,yuan2024satsense}. PGD\cite{madry2017towards} extends FGSM with an iterative process and projection step, generating stronger adversarial samples through multiple updates within constrained perturbations. Most current attack methods are built on PGD\cite{dong2018boosting,xie2019improving,lin2019nesterov}. Momentum Iterative MI-FGSM\cite{dong2018boosting} incorporates momentum and gradient iteration techniques to stabilize update directions. DI\textsuperscript{2}-FGSM\cite{xie2019improving} applies random transformations (e.g., scaling, translation, and rotation) to the input image each iteration before calculating gradients. NI-FGSM\cite{lin2019nesterov} introduces the Nesterov gradient acceleration technique to generate more effective adversarial samples and improve their transferability.

In DRL, Huang et al. \cite{huang2017adversarial} first use FGSM to study its ability to interfere with agent policies. Lin et al. \cite{lin2017tactics} propose an enchanting attack to maliciously manipulate DRL agents through strategic attacks on agents over a sequence of time steps. Weng et al.\cite{weng2019toward} present a model-based DRL attack framework designed to enhance the efficacy of adversarial attacks in DRL. Other works focus on adversarial defense, primarily using adversarial examples to train agents to improve policy robustness. Fischer et al.\cite{fischer2019online} devise Robust Student-DQN (RS-DQN), a method that enables simultaneous robust online training of Q-networks and policy networks. Zhang et al.\cite{zhang2020robust} proposed SA-MDP to theoretically unify the adversarial defense process in DRL. Oikarinen et al.\cite{oikarinen2021robust} develop RADIAL-RL framework to enhance the effectiveness of adversarial defense. 
However, these methods compute perturbations for attack or defense based on actions sampled from the policy (or actions with the highest weights), thus failing to capture broader vulnerability across the policy distribution.


\section{Methodology}
\subsection{Problem Formalization}
DRL is formulated as a Markov decision process\cite{sutton2018reinforcement}, which is formalized as a tuple $\left \langle \mathcal{S}, \mathcal{A},P,R,\gamma  \right \rangle $, where $\mathcal{S}$ and $\mathcal{A}$ denote the action and state spaces. Policy $\pi : \mathcal{S} \to \mathcal{P\left ( A \right ) } $ represents a probability distribution mapping from the state space to the action space. At each time step $t$, the agent chooses an action $a_t\sim \pi \left ( \cdot \mid s_t \right ) $ based on the current state $s_t$ and interacts with the environment. The environment transitions to a new state $s_{t+1}$ following the state-transition probability function $P\left ( s_t\mid s_{t+1},a_{t+1} \right ) $ and provides a reward $R\left ( s_t,a_t \right ) $ to the agent. The agent's goal is to learn a policy $\pi$ to maximize the expected discounted return $\mathbb{E} \left [  {\textstyle \sum_{i=0}^{\infty }\gamma ^iR\left ( s_{t+i} , a_{t+i}\right ) } \right ]$, where $\gamma$ denotes the discount factor. In an adversarial attack to policy, assuming $\pi _\theta \left ( \cdot \mid s \right ) $ represents a parameterized policy that has converged in the environment, where $\theta$ denotes the network parameters of the policy. For notational simplicity, we use $\pi \left ( \cdot \mid s \right )$ as $\pi \left [ s \right ]$. Given a state-action pair $\left( s , a \right )$, the goal of the adversarial attack is to find an adversarial example $s^*$ that maximizes the loss function while satisfying a paradigm constraint\cite{xu2020adversarial}:
\begin{equation}
\label{eq:1}
{\mathrm{arg} \max}_{s^*} J\left ( s^*,a \right ) ,\quad \mathrm{s.t.} \ \left \| s^*-s \right \| _p\le \epsilon, 
\tag{1}
\end{equation}
\noindent where $s^*=s+  \eta $ and $\eta$ denote the adversarial perturbation, $J\left ( s^*,a \right )$ represents the loss function, and $\left \| \cdot \right \| _p$ is the $L_p$ norm. In cases where the state and action space are discrete, the loss function is the cross-entropy loss, while for continuous cases, it is the mean square error loss. $L_p$ norm is usually the $L_1$ norm, $L_2$ norm or $L_\infty$ norm.

\subsection{Distribution Similarity Projected Gradient Descent}
\begin{algorithm}[t]
\small
\caption{DAPGD with $L_\infty$ norm constraint}
\label{alg:DAPGD}
\textbf{Input}: State $s$, stochastic policy $\pi_\theta \left [ s  \right ] $, iteration step $\alpha $, the number of iterations $N$, scaling factor $\varepsilon $, constraint threshold $\epsilon $. \\
\textbf{Output}: Adversarial example $s^*$. 
\begin{algorithmic}[1]
\STATE $s^* \gets  s+\varepsilon \cdot \mathcal{N} \left ( \mathbf{0},\mathbf{I}   \right )$
\FOR {$k=0$ to $N-1$}
\STATE Calculate the distribution similarity loss: 
\begin{equation*}
J \left ( \pi_\theta \left [ s^*  \right ],\pi_\theta \left [ s  \right ] \right ) = - \ln \int_{\mathcal{A} } \sqrt{ \pi_\theta \left ( a \mid s^* \right ) \pi_\theta \left ( a \mid s \right )} da
\end{equation*}
\STATE Calculate the gradient: 
\begin{equation*}
\mathbf{grad}\left [ s^*  \right ] =\bigtriangledown _{s^*}J \left ( \pi_\theta \left [ s^*  \right ],\pi_\theta \left [ s  \right ] \right )
\end{equation*}
\STATE $s^*\gets s^*+  \alpha \cdot \text{sgn}\left ( \mathbf{grad}\left [ s^*  \right ]  \right ) $
\STATE $s^*\gets s^*+  \text{clip}\left ( s^*- s,- \epsilon ,\epsilon   \right )$
\ENDFOR
\RETURN $s^*$
\end{algorithmic}
\end{algorithm}

The core idea of gradient-based attack methods is to calculate the gradient of the input data and then apply small perturbations to the input data along the gradient direction to induce incorrect output from the model\cite{goodfellow2014explaining,madry2017towards}. Numerous studies have demonstrated that this method can also significantly degrade the performance of DRL. For applying gradient-based attack methods to DRL, existing approaches typically rely on policy sampling to generate adversarial examples. To motivate this, we use the PGD\cite{madry2017towards} as an example to illustrate this process. Specifically, given a state-action pair $\left( s , a \right )$, PGD-based adversarial examples in DRL can be formulated as:
\begin{equation}
\label{eq:2}
\begin{aligned}
&& s^*_{k+1} & = s^*_k+  \alpha \cdot \text{sgn} \left ( \bigtriangledown _{s^*}J \left ( a^*_k,a \right )  \right )  \\
&& \text{where} \ \ s^*_0 & = s+\varepsilon \cdot \mathcal{N} \left ( \mathbf{0},\mathbf{I}   \right ) , \\
&& a^*_k & \sim \pi_\theta \left [ s^*_k  \right ], a \sim \pi_\theta \left [ s  \right ],  \\
&& k & = 0,1,\dots ,N-1,
\end{aligned}
\tag{2}
\end{equation}
\noindent where $\alpha$ denotes the iteration step size, $N$ is the number of iterations, $ \text{sgn}\left ( \cdot   \right )  
$ represents the sign function, $\varepsilon$ is the scaling factor, and $\mathcal{N} \left ( \mathbf{0},\mathbf{I}   \right )$ denotes the multivariate standard normal distribution. Although sampling actions to attack policies only calculate the sign gradient for one action in each iteration to generate adversarial examples. However, in continuous action spaces, agents can select other adjacent actions to counteract the impact of attacks targeting specific actions. This continuity provides a degree of fault tolerance and flexibility, allowing the policy to remain relatively stable under small perturbations. Moreover, in many practical applications, multiple near-optimal policies are common. For example, in navigation tasks, there are often several paths leading to the goal. This means that in the same state, indicating that multiple high-probability actions may exist for the same state. The sign gradient based on a single sampled action reflects only the local vulnerability of the policy, thereby limiting the effectiveness of the attack.

To overcome these limitations, we construct an optimization objective based on the similarity of policy distributions to generate adversarial examples. This design reduces the impact of randomness introduced by action sampling and leverages the complete probability information of the agent's action choice in a given state instead of individual action. Specifically, our objective is to maximize the following loss function under $L_p$ norm constraint:
\begin{equation}
\label{eq:3}
{\mathrm{arg} \max}_{s^*} J\left ( \pi_\theta \left [ s^*  \right ] ,\pi_\theta \left [ s  \right ] \right ) ,\ \mathrm{s.t.} \ \left \| s^*-s \right \| _p\le \epsilon. 
\tag{3}
\end{equation}

We utilize the Bhattacharyya distance\cite{kailath1967divergence} as a metric for measuring the similarity between policy distributions. Adversarial attacks fundamentally act as perturbations to the policy distribution, and the Bhattacharyya distance effectively quantifies the degree of overlap between two distributions. By maximizing the distance, we can induce an overall bias in the policy distribution. This approach is particularly effective in continuous action spaces,  as it simultaneously impacts high-probability actions and their adjacent alternatives, thereby exposing potential vulnerabilities in the policy. Specifically, we define the distribution similarity loss, built on the Bhattacharyya distance between policy distributions, as follows:
\begin{equation}
\label{eq:4}
J\left (  \pi_\theta \left [ s^*  \right ] ,\pi_\theta \left [ s  \right ] \right ) = - \ln \int_{\mathcal{A} } \sqrt{ \pi_\theta \left ( a \mid s^* \right ) \pi_\theta \left ( a \mid s \right )} da
\tag{4}
\end{equation}


Based on Eq.~\ref{eq:4}, the process of generating adversarial samples by DAPGD can be expressed as follows:

\begin{equation}
\label{eq:5}
\begin{aligned}
&& s^*_{k+1} & = s^*_k+  \alpha \cdot \text{sgn} \left ( \bigtriangledown _{s^*}J\left (  \pi_\theta \left [ s^*  \right ] ,\pi_\theta \left [ s  \right ]\right )  \right ) \\
&& \text{where} \ \ s^*_0 & = s+\varepsilon \cdot \mathcal{N} \left ( \mathbf{0},\mathbf{I}   \right ) , k = 0,1,\dots ,N-1.
\end{aligned}
\tag{5}
\end{equation}

The pseudocode for DAPGD is presented in Algorithm \ref{alg:DAPGD}. As there are no constraints on gradient computations or iteration methods, the distribution similarity loss is highly versatile and can be integrated with most existing adversarial attack methods. Moreover, it is applicable to DRL algorithms that utilize stochastic policies, whether in discrete or continuous action spaces.


\section{Evaluation}

\subsection{Experimental Setup}
\textit{Environments} We conduct extensive experiments on three continuous control navigation tasks from the Safety Gymnasium framework\cite{ji2023safety}: SafetyRacecarButton1-v0 (\textit{Button}), SafetyRacecarCircle1-v0 (\textit{Circle}), and SafetyRacecarGoal1-v0 (\textit{Goal}). These tasks simulate scenarios relevant to autonomous vehicle navigation, such as activating buttons, maintaining circular trajectories, and reaching targets, while adhering to safety constraints like avoiding hazards and boundaries. The agent operates with realistic car dynamics in a 2-dimensional action space (velocity and steering angle), and the observation space includes multiple sensor data from lidar, accelerometer, speedometer, gyroscope, and magnetometer.

\textit{Agent and Training Configuration} 
We evaluate the effectiveness of various adversarial attack methods on DRL agents trained for continuous control navigation tasks. In each environment, agents are trained using the Trust Region Policy Optimization (TRPO) algorithm\cite{schulman2015trust}. The agents are configured with 20 conjugate gradient iterations, 15 search steps, 10 learning iterations, a discount factor $\gamma$ of 0.99, a batch size of 128, and a maximum gradient norm of 50, trained for $1 \times 10^7$ steps. All networks consist of two hidden layers with 64 nodes.

\textit{Attack Methods} We implement and evaluate a range of attack methods, including FGSM\cite{goodfellow2014explaining}, DI\textsuperscript{2}-FGSM\cite{xie2019improving}, MI-FGSM\cite{dong2018boosting}, NI-FGSM\cite{lin2019nesterov}, PGD\cite{madry2017towards}, TPGD\cite{zhang2019theoretically}, EOTPGD\cite{liu2018adv}, and our proposed DAPGD method. For iterative methods, we conduct experiments with the number of
iteration $N=50 \ \text{or} \ 100$. We set the iteration step $\alpha = 2/255$,  scaling factor $\varepsilon = 0.001$, and constraint threshold $\epsilon =0.1$.

\textit{Evaluation Schemes} Two schemes are used for performance evaluation: a) Direct attacks: Attack the models trained without noise. b) Post-defense attacks: Attack models that have undergone additional defensive training with PGD (the number of iteration $N=50$) for $5 \times 10^6$ training steps, keeping other parameters unchanged. This setup demonstrates DAPGD's effectiveness on both benign DNNs (without adversarial training) and robust DNNs (with adversarial training). The
agents’ performance under various attack methods is measured by the average episode reward obtained over 1000 episodes in three independent experiments. Attacks are applied to the agents’ observation, perturbing the input before it is processed by the policy network. We record the mean reward obtained
by the agents under each attack scenario, with lower rewards indicating more successful attacks.

\subsection{Overall performance of DAPGD }

In Table~\ref{t1}, all attack methods reduce the agent's average reward across the three tasks (Goal, Circle, Button) compared to the non-attacked case. The DAPGD method demonstrates the best attack performance across all tasks. DAPGD achieves an average reward reduction 18.67\% higher than the best baseline.

Table~\ref{t2} presents the attack performance under the defensed model (after adversarial training with PGD). While the agent exhibits a certain degree of robustness following defense training, DAPGD still causes a significant performance drop. DAPGD outperform all other methods across all tasks, causing the most significant reduction in rewards, average 25.38\% higher than the best baseline. In the presence of defense mechanisms, DAPGD's advantage over baselines becomes even more pronounced. This is attributed to DAPGD more effectively attacking the policy distribution rather than relying solely on sampling individual actions, thus revealing vulnerabilities that PGD struggle to defend against.


\begin{table}[]
\renewcommand{\arraystretch}{1.2}
\caption{Average reward in direct attack setting}
\begin{tabular}{lllll}
\cline{3-5}
\multicolumn{2}{c}{}                                  & \multicolumn{3}{c}{Task}                                                         \\ \hline
\multicolumn{1}{c}{Method} & \multicolumn{1}{c}{Iters} & \multicolumn{1}{c}{Goal} & \multicolumn{1}{c}{Circle} & \multicolumn{1}{c}{Button} \\ \hline
No attack                  & -                        & 26.043±0.223             & 42.144±0.023               & 9.148±0.343                \\
FGSM                       & -                        & 23.059±0.252             & 40.889±0.013               & 7.478±0.804                \\
DI\textsuperscript{2}-FGSM                     & 50                       & 21.360±0.338             & 39.850±0.028               & 7.900±0.711                \\
DI\textsuperscript{2}-FGSM                     & 100                      & 20.807±0.603             & 39.931±0.038               & 7.414±1.047                \\
MI-FGSM                     & 50                       & 21.689±0.328             & 40.115±0.049               & 7.160±0.797                \\
MI-FGSM                     & 100                      & 20.929±0.788             & 40.075±0.008               & 7.339±0.448                \\
NI-FGSM                     & 50                       & 23.630±0.694             & 38.769±0.048               & 7.143±0.497                \\
NI-FGSM                     & 100                      & 24.626±0.185             & 37.187±0.121               & 7.274±0.716                \\
PGD                        & 50                        & 24.642±0.361             & 41.391±0.055               & 8.445±0.433                \\
PGD                        & 100                        & 22.958±0.602             & 40.890±0.055               & 6.973±0.450                \\
TPGD                       & 50                       & 20.336±1.314             & 33.870±0.079               & 6.546±0.815                \\
TPGD                       & 100                      & 20.651±0.512             & 33.752±0.099               & 6.952±0.581                \\
EOTPGD                     & 50                       & 21.031±0.996             & 40.272±0.007               & 7.097±0.642                \\
EOTPGD                     & 100                      & 20.982±1.236             & 40.243±0.012               & 6.775±0.105                \\
DAPGD (ours)               & 50                       & 20.074±0.446             & \textbf{33.351±0.043}      & 5.934±0.764                \\
DAPGD (ours)               & 100                      & \textbf{19.965±0.813}    & 33.467±0.082               & \textbf{5.382±0.310}       \\ \hline
\end{tabular}
\vspace{5pt} 
\label{t1}
\end{table}

\begin{table}[]
\renewcommand{\arraystretch}{1.2}
\caption{Average reward in post-defense attacks setting (defended by PGD, the number of iteration $N=50$) }
\begin{tabular}{lllll}
\cline{3-5}
\multicolumn{2}{c}{}                                  & \multicolumn{3}{c}{Task}                                                         \\ \hline
\multicolumn{1}{c}{Method} & \multicolumn{1}{c}{Iters} & \multicolumn{1}{c}{Goal} & \multicolumn{1}{c}{Circle} & \multicolumn{1}{c}{Button} \\ \hline
PGD (defended)                     & 50                        & 24.746±0.574             & 39.855±0.098               & 9.291±1.079                \\ 
PGD                      & 100                        & 22.354±0.221             & 39.461±0.019               & 8.995±0.658                \\ 
FGSM                       & -                        & 23.852±0.320             & 38.461±0.055               & 9.098±0.402                \\
DI\textsuperscript{2}-FGSM                     & 50                       & 23.408±0.465             & 40.402±0.022               & 8.656±0.975                \\
DI\textsuperscript{2}-FGSM                     & 100                      & 22.471±0.278             & 40.400±0.050               & 7.873±0.168                \\
MI-FGSM                     & 50                       & 23.336±0.435             & 39.119±0.066               & 8.326±0.146                \\
MI-FGSM                     & 100                      & 22.741±0.456             & 39.207±0.018               & 9.109±0.513                \\
NI-FGSM                     & 50                       & 24.380±0.472             & 39.988±0.064               & 8.444±0.485                \\
NI-FGSM                     & 100                      & 24.883±0.453             & 40.369±0.046               & 8.890±0.845                \\
TPGD                       & 50                       & 22.456±0.855             & 38.114±0.038               & 8.422±0.195                \\
TPGD                       & 100                      & 22.739±0.378             & 38.038±0.025               & 8.001±0.388                \\
EOTPGD                     & 50                       & 22.990±0.273             & 39.326±0.046               & 7.825±0.325                \\
EOTPGD                     & 100                      & 22.948±0.245             & 39.417±0.069               & 8.337±0.778                \\
DAPGD (ours)               & 50                         & 22.701±0.438             & 38.028±0.057               & \textbf{7.511±0.347}       \\
DAPGD (ours)               & 100                         & \textbf{21.179±0.151}    & \textbf{37.936±0.085}      & 8.033±0.693                \\ \hline
\end{tabular}
\vspace{5pt} 
\label{t2}
\end{table}

\begin{figure}
    \centering
    \includegraphics[width=0.7\linewidth]{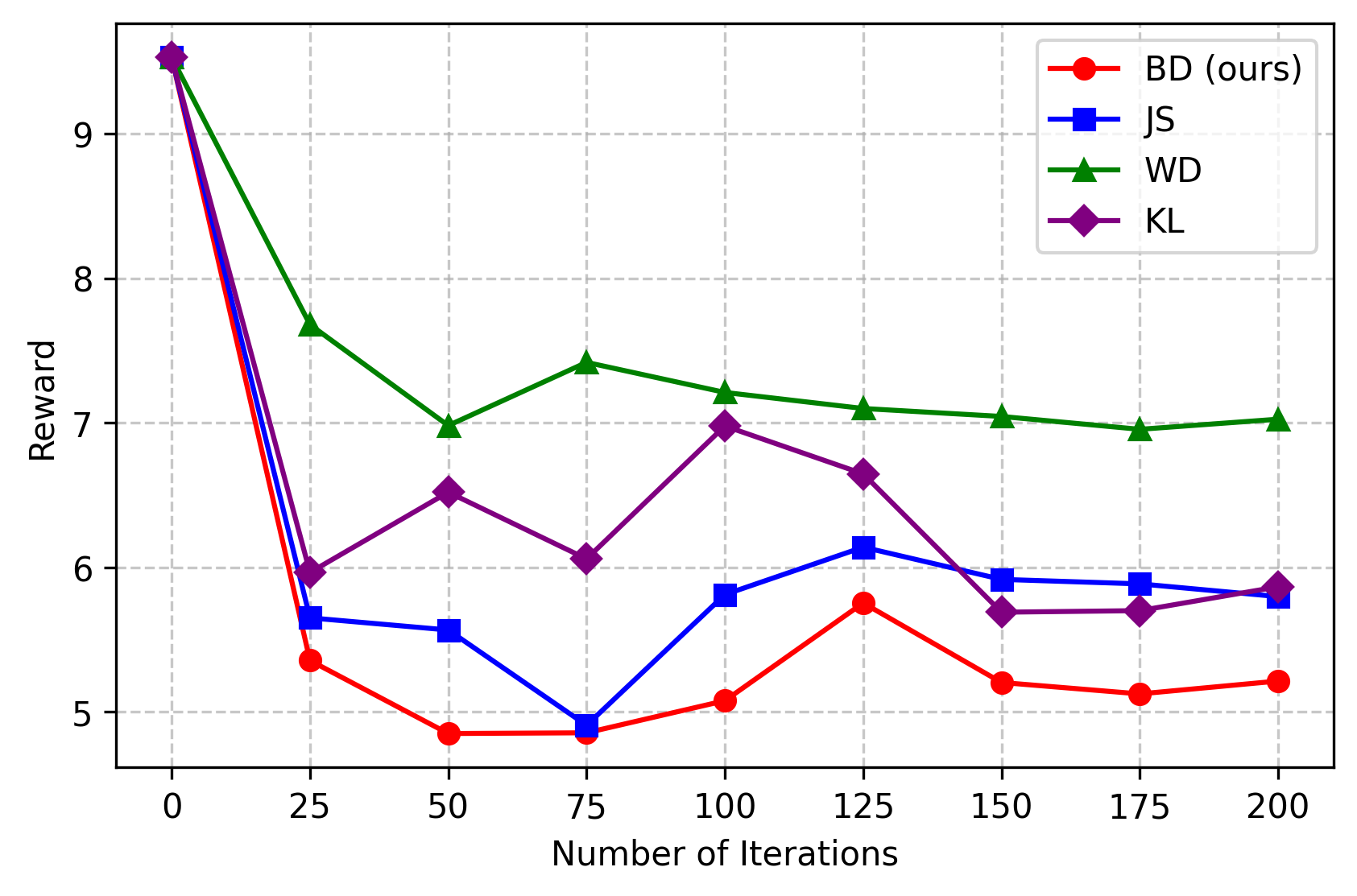}
    \caption{Average reward obtained by the agent under each attack configuration in \textit{Button}. Lower rewards indicate more effective attacks.}
    \label{fig:ablation}
\end{figure}

To further investigate the effectiveness of the use of Bhattacharyya distance (BD) in DAPGD, we conducted an ablation study comparing the following three alternative metrics Kullback-Leibler Divergence (KL), Jensen-Shannon Divergence (JS), and 2-Wasserstein Distance (WD)\cite{ruschendorf1985wasserstein} used in the attack. Figure~\ref{fig:ablation} illustrates the performance of different divergence metrics across different numbers of iterations of attack. BD achieves the lowest reward values across all iteration counts, indicating its ability to craft more potent attacks. The superior performance of BD can be attributed to its comprehensive capture of distribution disparities. By maximizing the BD,  a broader shift in the policy distribution is induced, impacting both high-probability actions and their neighbors.

\section{Conclusion}
We propose DAPGD, a novel adversarial attack method for DRL agents. DAPGD utilizes distribution similarity and targets the entire policy distribution addressing limitations of existing attacks, especially in continuous action spaces. It use Bhattacharyya distance to measure policy
similarity, enabling sensitive detection of subtle but critical differences between probability distributions.  Experimental results show that DAPGD outperforms the all baselines in three navigation tasks, achieving average 22.03\% higher reward drop compared to the best baseline. As a potential future direction, we are looking forward to extending our DAPGD to improve the performance of various applications such as large language models~\cite{lin2024splitlora,hu2021lora,fang2024automated}, multi-modal training~\cite{zheng2023autofed,hu2024t,fang2024ic3m,tang2024merit}, distributed machine learning~\cite{zw2024specbreathing,leo2025split} and autonomous driving~\cite{hu2024agentscomerge,lin2022tracking,zheng2023autofed,hu2024agentscodriver,lin2022channel}. 
\bibliographystyle{IEEEtran}
\bibliography{reference}
\end{document}